%% file: main.tex
\pdfoutput=1

\PassOptionsToPackage{table}{xcolor}
\PassOptionsToPackage{dvipsnames}{xcolor}

\documentclass[11pt]{article}

\usepackage[preprint]{acl}

\usepackage{booktabs}
\usepackage{times}
\usepackage{latexsym}
\usepackage{booktabs}
\usepackage{array}
\usepackage{amsmath}
\usepackage{mathabx}
\usepackage{multirow}
\usepackage{multicol}

\usepackage[T5]{fontenc}

\usepackage[utf8]{inputenc}
\usepackage{enumitem} 

\usepackage{microtype}

\usepackage{inconsolata}

\usepackage{graphicx}
\usepackage{xcolor}
\usepackage[dvipsnames]{xcolor}

\usepackage{mathrsfs}
\usepackage{lipsum} 
\usepackage[normalem]{ulem}
\usepackage{wrapfig}
\usepackage{float}
\usepackage{tcolorbox}
\usepackage{color, colortbl}
\usepackage{tcolorbox}
\usepackage{xcolor}
\usepackage{soul}
\usepackage{subcaption}
\usepackage{caption}

\newcommand{\colorhl}[2]{\sethlcolor{#1}\hl{#2}}

\DeclareRobustCommand{\github}{%
  \begingroup\normalfont
  \vspace{0.5em}%
  \raisebox{-0.3em}{%
  \includegraphics[height=1.3em]{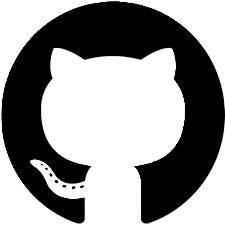}%
  }%
  \kern 0.4em%
  \endgroup
}

\DeclareRobustCommand{\mail}{%
  \begingroup\normalfont
  \vspace{0em}%
  \raisebox{0em}{%
  \includegraphics[height=0.8em]{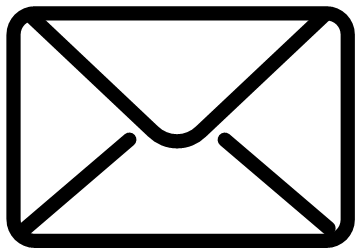}%
  }%
  \kern 0.4em%
  \endgroup
}

\definecolor{custom_light_blue}{rgb}{0.85, 0.95, 1}
\definecolor{custom_light_pink}{rgb}{1, 0.85, 0.85}
\definecolor{custom_light_green}{rgb}{0.85, 0.98, 0.80}
\newcommand{\sentimentreasoning}{\texttt{\colorhl{blue!15}{Sentiment Reasoning}}}
\newcommand{\rationalegeneration}{\texttt{\colorhl{BrickRed!15}{Rationale Generation}}}
\newcommand{\sentimentclassification}{\texttt{\colorhl{yellow!50}{Sentiment Classification}}}

\title{Sentiment Reasoning for Healthcare}
\author{Khai-Nguyen Nguyen$^{*1}$, Khai Le-Duc$^{*2,3}$,  \\ \bf{Bach Phan Tat$^{4}$,  Duy Le$^{5}$, Long Vo-Dang$^{6}$, Truong-Son Hy$^{7}$}\\
$^1$College of William and Mary, USA
$^2$University of Toronto, Canada\\
$^3$University Health Network, Canada
$^4$KU Leuven, Belgium \\
$^5$Bucknell University, USA
$^6$University of Cincinnati, USA \\
$^7$University of Alabama at Birmingham, USA\\
\mail \texttt{knguyen07@wm.edu \hspace{0.5cm}} \mail \texttt{duckhai.le@mail.utoronto.ca} \\
\github \colorbox{custom_light_green}{\texttt{\href{https://github.com/leduckhai/Sentiment-Reasoning}{https://github.com/leduckhai/Sentiment-Reasoning}}}}

\begin{document}
\maketitle
\begin{abstract}
Transparency in AI healthcare decision-making is crucial. By incorporating rationales to explain reason for each predicted label, users could understand Large Language Models (LLMs)’s reasoning to make better decision. In this work, we introduce a new task - \sentimentreasoning\ - for both speech and text modalities, and our proposed multimodal multitask framework and  {\textbf{the world's largest multimodal sentiment analysis dataset}}. \sentimentreasoning\ is an auxiliary task in sentiment analysis where the model predicts both the sentiment label and generates the rationale behind it based on the input transcript. Our study conducted on both human transcripts and Automatic Speech Recognition (ASR) transcripts shows that \sentimentreasoning\  helps improve model transparency by providing rationale for model prediction with quality semantically comparable to humans while also improving model's classification performance (\textbf{{+2\% increase in both accuracy and macro-F1}})  via rationale-augmented fine-tuning. Also, no significant difference in the semantic quality of generated rationales between human and ASR transcripts. All code, data (five languages - Vietnamese, English, Chinese, German, and French) and models are published online.

\end{abstract}

\def\thefootnote{(*)}\footnotetext{Equal contribution}\def\thefootnote{\arabic{footnote}}

\section{Introduction}

\begin{figure*}
    \centering
    \includegraphics[width=\linewidth]{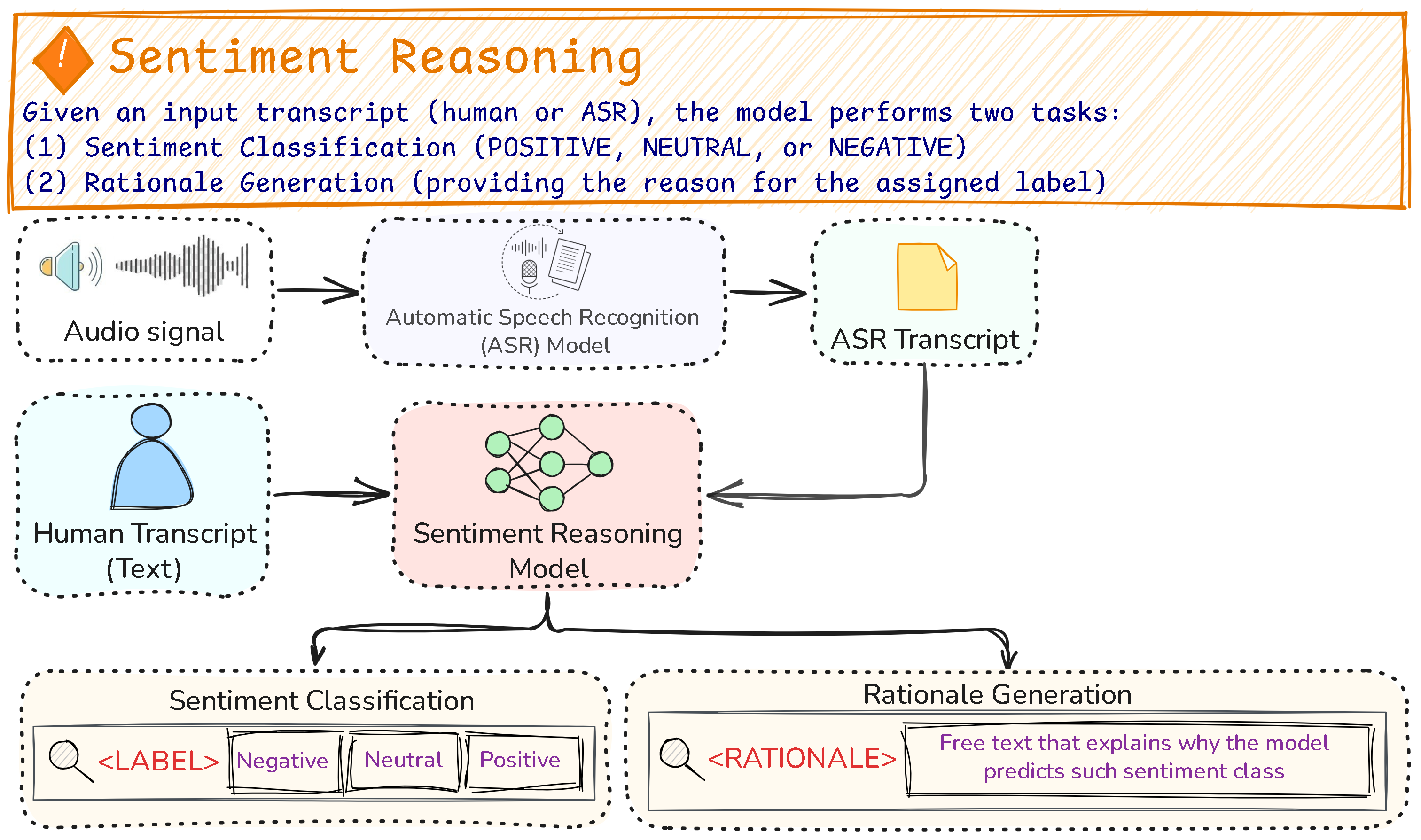}
    \caption{Visualized pipeline for \sentimentreasoning\  . Given an input transcript (either human transcript or ASR transcript), the model learns to output the \textbf{sentiment label} (POSITIVE, NEUTRAL, or NEGATIVE) and its \textbf{rationale} (the reason for this label). It comprises of two tasks: (1) \sentimentclassification\ and (2) \rationalegeneration\ . Traditional sentiment analysis only includes \sentimentclassification\ task, while our framework generates corresponding rationale to explain the reason behind each predicted sentiment label. 9 examples with sentiment labels and their corresponding rationales in our dataset are shown in Table \ref{data_samples} in the Appendix.}
    \label{fig:sentiment_reasoning_pipeline}
\end{figure*}

Sentiment analysis plays a pivotal role within the healthcare domain. In healthcare customer service, it facilitates real-time evaluation of customer satisfaction, enhancing empathetic and responsive interactions \cite{xia2009improving, na2012sentiment}. Moreover, sentiment analysis aids in monitoring the emotional well-being of patients \cite{cambria2012sentic}, including those with mental health issues such as suicide \cite{pestian2012sentiment}. However, these studies only work on text-only sentiment analysis instead of speech-based sentiment analysis.

Despite its potential, speech sentiment analysis presents several technical challenges. First, emotions conveyed through speech are subjective \cite{wearne2019emotion}, complex \cite{golan2006reading}, and dependent on speaking styles \cite{shafran2003robust}, making accurate sentiment classification difficult even for humans \cite{kuusikko2009emotion}, thereby necessitating the role of explainable artificial intelligence (AI). Second, given the critical nature of healthcare decisions, where errors can have severe consequences, transparency in AI decision-making is essential to build trust among machines, healthcare professionals, and patients \cite{antoniadi2021current}.

To tackle challenges above, reasoning in AI is crucial for sentiment analysis because it enables deeper understanding beyond surface-level sentiment polarity via the textual explanations. 
Recent works on Chain-of-Thought (CoT) distillation \cite{wadhwa2024investigating, chen2024post, hsieh2023distilling, ho2022large} have revealed that training generative small language models (SLMs) on rationale-augmented targets (the CoT from larger models is provided along side with the target label) can help the SLM (1) perform better and (2) acquire the ability to generate rationale. Our work leverage these findings and prepare a set of human-labeled rationale to train our sentiment analysis models to do \rationalegeneration\ and enhance their performance (Section \ref{trainingwithrationale} and \ref{sec:rationale_format}).
By incorporating rationales to explain reason for each predicted sentiment label, users could understand the model's reasoning, facilitating better decision-making based on the classification results. Therefore, we introduce a novel multimodal framework for a novel task: \sentimentreasoning, which comprises of two tasks: (i) \sentimentclassification, in which the model learns to output the \textbf{sentiment label} (POSITIVE, NEUTRAL, or NEGATIVE), and (ii) \rationalegeneration, in which the model generates rationale (the free-form text that explains reason for this label). Our contributions are as follows:

\begin{enumerate}
[noitemsep,topsep=0pt,parsep=0pt,partopsep=0pt]
    \item We introduce a new task: \sentimentreasoning\   for both speech and text modalities, along with the world's largest multimodal sentiment analysis dataset, supporting five languages (Vietnamese, English, Chinese, German, and French)
    \item We propose our novel multimodal speech-text \sentimentreasoning\   framework
    \item We empirically evaluate the baselines on our dataset using state-of-the-art backbone models
    \item We provide in-depth analysis of rationale / Chain-of-Thought (CoT)-augmented training 
\end{enumerate}

All code, data and models are published online.

\section{Data}

\subsection{Data Collection}
The dataset employed for constructing the \sentimentreasoning\   dataset was \textit{VietMed} \cite{vietmed_dataset}, a large and publicly accessible medical ASR dataset. The dataset comprises real-world doctor-patient conversations. We then annotated \textit{sentiment labels} (POSITIVE, NEUTRAL, or NEGATIVE) and their corresponding \textit{rationales} (the reason for this label). We then manually translate the transcripts from Vietnamese into other four languages: English, Chinese (Simplified and Traditional), German, and French, making the dataset six times larger.
The full dataset (with 5 languages) includes 30000 samples, making it \textbf{the largest multimodal sentiment analysis dataset}, to the best of our knowledge (see Table \ref{tab:datastats_literature}). Our paper focuses mainly on the \textbf{Vietnamese subset} (Section \ref{sec:results_and_analysis}) and the \textbf{English subset} (Appendix \ref{sec:results_on_English_subset}). 


\subsection{Data Annotation}
The annotation task consists of two primary steps. First, annotators are required to perform \sentimentclassification . Second, annotators are instructed to provide a rationale behind each class (\rationalegeneration\ ). To ensure consistency, our TESOL-certificated professional linguist has developed an initial guideline inspired by \cite{switchboard_sentiment}, which was also adopted by various well-known works \cite{shon2022slue_dataset, shon2023slue}, and revised it frequently if necessary. Details of data annotation pipeline, annotation guidelines, data imbalance, translation annotation, and translaton quality control are shown in Appendix Section \ref{sec:Appendix_data}.

\subsection{Data Quality Control}
During the independent annotation process conducted by three annotators, we observed a low inter-annotator agreement (Cohen’s kappa coefficient below 0.5 for the inter-annotator agreement between the two annotators), a common occurrence in real-world datasets as noted by \citet{switchboard_sentiment}. To address this issue, we implemented an alternative label merging approach. We convened a discussion meeting involving the three annotators and two reviewers (one professional linguist and one with a biomedical background). Each annotator was required to justify their chosen sentiment label and its corresponding rationale. A label and its rationale were selected based on the consensus of all three annotators and two reviewers, rather than a majority vote, as employed in other studies \cite{aziz2020twitter, saleena2018ensemble}.

\subsection{Data Statistics}
\input{tables_and_figures/label_distribution}
\input{tables_and_figures/datastats_literature}

Table \ref{tab:label_distribution} shows the distribution of sentiment labels in the dataset.
This reflects the dataset's slight emphasis on neutral content, typical in medical conversations involving explanations and advice. 

It should be noted that the statistics are reported \textbf{for a single language}, meaning that the real size of the dataset is 6 times larger when accounting all 5 languages.

\section{Sentiment Reasoning Framework}

\subsection{Informal Definition}
As shown in Figure \ref{fig:sentiment_reasoning_pipeline}, in \sentimentreasoning\  , given an input transcript (either human transcript or ASR transcript), the model learns to output the \textbf{sentiment label} (POSITIVE, NEUTRAL, or NEGATIVE) and its \textbf{rationale} (the reason for this label). It comprises of two tasks: \sentimentclassification\ and \rationalegeneration\ . 


\subsection{Formal Definition}
Let $x^{T}_{1} := x_{1}, x_{2}, ..., x_{T}$ be an audio signal of length $T$. Let $C$ be the set of all possible sentiment classes, we should build a speech-based \sentimentreasoning\   model $f$ that both estimates the probability $p(c|x^{T}_{1})$ for each $c \in C$ and generates its rationale sequence $r^{M}_{1}$ of $M$ length. 

The decision rule to predict a sentiment class is: 
\begin{equation}
x^{T}_{1} \to \hat{c} = \arg\max_{c \in C} f(c|x^{T}_{1})
\end{equation}

The decision rule to generates the corresponding rationale sequence is: 
\begin{equation}
x^{T}_{1} \to r^{M}_{1} = \arg\max_{r^{*}} h(r^{*}|x^{T}_{1})
\end{equation}

For text-based \sentimentreasoning, the input audio signal $x^{T}_{1}$ could be replaced with a word sequence (human transcript) $w^{N}_{1}$ of length N, thus ASR model is not needed.

\subsection{ASR Model}
An ASR model aims to convert audio signal into text by mapping an audio signal $x^{T}_{1}$ to the most likely word sequence $w^{N}_{1}$. 
The relation $w^{*}$ between the acoustic and word sequence is:
\begin{equation}
w^{*} = \operatorname{arg}\max_{w_1^N} \, p(w_{1}^{N}|x_{1}^{T})    
\end{equation}

\subsection{Language Model for Sentiment Reasoning}
\subsubsection{Sentiment Classification}
Let the transcribed audio signal (ASR transcript) $w^{N}_{1}$ serve as the input for the \sentimentclassification\ model $g$, which maps $w^{N}_{1}$ to a class label $\hat{c}$:
\begin{equation}
w^{N}_{1} \to \hat{c} = \arg\max_{c \in C} g(c|w^{N}_{1})
\end{equation}

$g$ is trained to minimize a loss function $\mathscr{L}(g(w^{N}_{1}), \hat{c})$. The optimal parameters $\theta$ of the model are found by solving the optimization problem $\min_{\theta} \mathscr{L}(g(w^{N}_{1}; \theta), \hat{c})$. Once trained, the model can predict the class of the transcribed audio signal by evaluating $\hat{c} = g(w^{N}_{1})$.

\subsubsection{Rationale Generation} 
Let the transcribed audio signal (ASR transcript) $w^{N}_{1}$ serve as the input for the \rationalegeneration\ model $h$, which maps $w^{N}_{1}$ to a rationale sequence $r^{M}_{1}$ of $M$ length:
\begin{equation}
w^{N}_{1} \to r^{M}_{1} = \arg\max_{r^{*}} h(r^{*}|w^{N}_{1})
\end{equation}

$h$ is trained to minimize a loss function $\mathscr{L}(h(w^{N}_{1}), r^{M}_{1})$. The optimal parameters $\theta$ of the model are found by solving the optimization problem $\min_{\theta} \mathscr{L}(g(w^{N}_{1}; \theta), r^{M}_{1})$. Once trained, the model can generate rationale of the transcribed audio signal by evaluating $r^{M}_{1} = h(w^{N}_{1})$.


\section{Experimental Setups}
\subsection{ASR Model}
We employed hybrid ASR setup using wav2vec 2.0 encoder \cite{vietmed_dataset} to transcribe speech to text. The final ASR model has 118M trainable parameters and Word-Error-Rate (WER) of 29.6\% on the test set. Details of ASR experiments are shown in Appendix \ref{sec:details_ASR_experiments}.

\subsection{End-to-end Sentiment Classification}
\textbf{PhoWhisper} \cite{PhoWhisper}: The encoder-based Whisper \cite{radford2022whisper} trained on an 844-hour Vietnamese dataset. To perform \sentimentclassification , we attach a classification head to the encoder layer of the model. We use the base version in our experiments.

\subsection{End-to-end Sentiment Reasoning}
\textbf{Qwen2-Audio} \cite{Qwen2-Audio}: We fine-tune the Qwen2-Audio (7B parameters) on both \textit{Label Only} and \textit{Label + Rationale} settings. We use the \textit{Instruct} version in our experiments.

\subsection{Language Model for Sentiment Reasoning}

\subsubsection{Encoder}
The encoder architecture is naturally well-suited for \sentimentclassification , which can be reformulated into the classical classification task. To this end, we directly apply a linear classifier to the output of the encoders. However, encoders can not generate rationales. As such, \textbf{they serve as baselines in our experiments}.

We use \textbf{phoBERT} (110M params) \cite{nguyen2020phobert}, RoBERTa \cite{liu2019roberta} pre-trained on 20GB Vietnamese text, and \textbf{ViHealthBERT} (110M params) \cite{minh2022vihealthbert}, phoBERT trained on 32GB of Vietnamese text in the healthcare domain. For ViHealthBERT, we report the syllable version which achieved better performance than the word version.

\subsubsection{Generative Models}
We reformulated \sentimentclassification\ into a text-to-text problem, where given the input transcript $w^{N}_{1}$, the generative model $g$ and the predicted sentiment class $c$, we have \textit{$g(w^{N}_{1})=c$} with $c \in C = \{\textit{"0", "1", "2"}\}$ where $"0", "1", "2"$ corresponds to the labels \textit{NEGATIVE}, \textit{NEUTRAL} and \textit{POSITIVE}. 

\textbf{Encoder-Decoder}:  \textbf{BARTpho} (139M params) \cite{tran2022bartpho} is the Vietnamese variant of BART \cite{lewis2019bart} trained on 20GB of Vietnamese text from Wikipedia and news corpus. \textbf{ViT5} (223M params) \cite{phan2022vit5} is the Vietnamese version of T5 \cite{raffel2020exploring} trained on 71GB of Vietnamese text from CC100 \cite{conneau2019unsupervised}.

\textbf{Decoder}: We use \textbf{Vistral-7B-Chat} \cite{chien2023vistral} and \textbf{vmlu-llm}\footnote{https://huggingface.co/vtrungnhan9/vmlu-llm}. Both models have Mistral-7B\cite{jiang2023mistral} as their backbone. These models were chosen based on their performance on the \textbf{vmlu benchmark} (\textbf{V}ietnamese \textbf{M}ultitask \textbf{L}anguage \textbf{U}nderstanding)\footnote{https://vmlu.ai/leaderboard}.

\subsection{Training with Rationale}
\label{trainingwithrationale}
Previous works \cite{wadhwa2024investigating, chen2024post,hsieh2023distilling,ho2022large} have shown that rationale-augmented targets consistently improve the performance of generative language models. Our rationale-augmented training methods are based on, to our knowledge, the current state-of-the-art CoT-distillation approaches for each architecture.

(i) \textbf{Multitask Training} \cite{hsieh2023distilling}: We train our encoder-decoders using distilling step-by-step. Distilling step-by-step is a multitask training approach that prepends particular prefixes to the input, guiding the model to output either the answer or generate a rationale. \citeauthor{hsieh2023distilling} found that it consistently improves encoder-decoders performance compared with single-task training which treats rationale and label predictions as a single task.

(ii) \textbf{Post-thinking} \cite{chen2024post}: For decoder-based models, we augment the training targets by append the human rationale to the label (<LABEL> <RATIONALE>) in a single prompt. Previous works have shown that post-thinking achieved impressive performance \cite{chen2024post, wadhwa2024investigating} and compared to pre-thinking where the model first generates its CoT then provide the label (<RATIONALE> <LABEL>), post-thinking is more stable and token-efficient \cite{chen2024post,wadhwa2024investigating} as the model suffers less from hallucination, consistently yields better performance and is more resource efficient as users can already retrieve the target label from the first generated token.

\subsection{Rationale Format}
\label{sec:rationale_format} 
While the rationale in our dataset were re-labeled by humans, we are also interested in \textbf{whether a different and more detailed rationale format would help the models learn better}.
To this end, we further study the effects of the format of the rationale on the performance of the generative models. In particular, given the human rationale and human label, we further prompt GPT-3.5-turbo to enhance the rationale into two different format:

    \noindent  \textbf{Elaborated rationale}: An elaborated version of the human rationale that is 1-2 sentence(s) long, grounded on the provided human rationale and the sentiment label.
    
    \noindent \textbf{CoT rationale}: A step-by-step, elaborated version of the human rationale, which includes the following steps: (1) identifies the medical entity, (2) extracts the progress of the corresponding medical entity in the transcript, and (3) provides the elaborated rationale on the sentiment grounded on the provided human rationale, the sentiment label, and information from steps (1) and (2). This approach is inspired by aspect-based sentiment instruction-tuning approaches \cite{varia2022instruction}.

\subsection{Evaluation Metrics}
For \sentimentclassification\ task, we employ accuracy and class-wise F1 score. For \rationalegeneration\ , we employ ROUGE (Recall-Oriented Understudy for Gisting Evaluation) score \cite{lin2004rouge}.
Also, we employ BERTScore  \cite{zhangbertscore}
which captures the contextual and semantic nuances. 
BERTscore has shown to correlate well with human judgment.

\section{Results and Analysis}
\label{sec:results_and_analysis}

\input{tables_and_figures/vi_humantranscript}
\input{tables_and_figures/vi_ASRtranscript}

\input{tables_and_figures/CoT_humantranscript}

\begin{figure}[!h]
  \centering
  \includegraphics[width=1.0\linewidth]{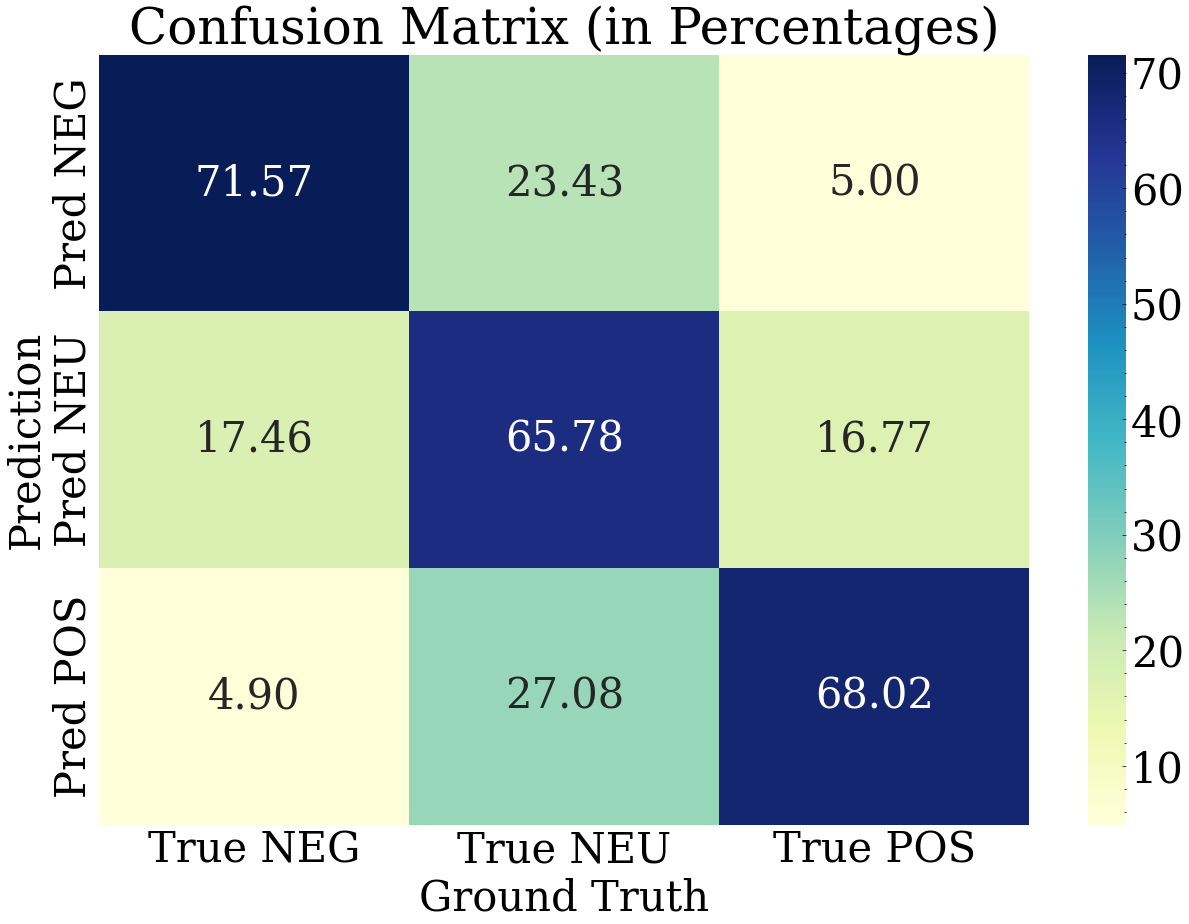}
  \caption{Confusion matrix of the predicted classes versus the actual labels on human transcript, obtained from Vistral7B trained with human rationale}
  \label{fig:jerryconfusion_matrix}
\end{figure}

We evaluate and analyze our models performance on Table \ref{vi_humantranscript}. Based on the obtained results, we make the following observations: 

\noindent \textbf{1. Encoders are efficient yet effective \sentimentclassification\ baselines}: Encoder models yields the best performance compared to their encoder-decoder and decoder counterparts, with high accuracy scores (> 0.665) and stable F1 scores (macro F1 of both models > 0.665). 
We further observe that \textbf{domain-specific encoders yield notably better performance}, with ViHealthBERT outperforming phoBERT in accuracy (+0.8\%) and macro F1 (+0.9\%).

\noindent \textbf{2. ASR errors have a marginally negative impact on \sentimentclassification\ performance}: For a fair comparison in real-world environments, WERs for human annotators on a standard conversational spontaneous English ASR dataset range from 5\% to 15\% \cite{stolcke2017comparing} while more challenging real-world ASR datasets are between 17\% and 31\% \cite{mulholland2016comparison}. 
Given the complexity of real-world medical conversations, WER of 29.6\% by our ASR model is within an acceptable range. 
Despite the WER of 29.6\%, the performance drop in macro F1 scores is small (absolute value of only about 5\%).

\noindent \textbf{3. Rationale-augmented training improve model performance:} Consistent with previous findings, performing CoT-augmented training on both encoder-decoders and decoders improve our models performance compared to the baseline. 
 
We further conducted a Student's t-test \cite{student1908probable} and found that the gains are statistically significant for $\alpha=0.1$. This pattern holds for the results in Table \ref{CoT_humantranscript}. We observe a decline in all of our models performance on ASR data which is anticipated due to its WER of 29.6 \%. Nonetheless, the models trained with rationale perform noticeably better than models without, with an average absolute accuracy gain of +0.85\%, absolute macro F1 gain of +1.4\%, and relative macro F1 gain of +2.5\%. 

\noindent \textbf{4. The format of post-thinking rationale doesn't affect the generative models performance}: We study the effects of the format of post-thinking rationale on the performance of generative models on Table \ref{CoT_humantranscript} and observe that it is unclear whether there is a performance gain from more elaborated rationales. This result agrees with previous findings \cite{wadhwa2024investigating}. 

\noindent \textbf{5. Models are likely to misclassify \textit{POSITIVE} and \textit{NEGATIVE} transcripts as \textit{NEUTRAL}}: We study the confusion matrix of our best model on human transcript, Vistral7B finetuned with human rationale, on Figure \ref{fig:jerryconfusion_matrix}. We observe a notable  misclassification tendency between \textit{NEUTRAL} and the other two classes (23.43\% and 27.08\% with \textit{NEGATIVE} and \textit{POSITIVE} respectively). On the other hand, we found that models can easily distinguish \textit{NEGATIVE} transcripts from \textit{POSITIVE} ones. This reflects the ambiguity of sentiment analysis data.
 
Furthermore, given the slightly imbalanced nature of our dataset with fewer \textit{POSITIVE} examples, its average F1 score is the lowest among the three labels across all models. 

\noindent \textbf{6. Analysis of Generated Rationale}:  
Compared to human rationale, we observe from Table \ref{vi_humantranscript} and Table \ref{vi_ASRtranscript} that the models trained with rationale have high BERTscore (around 0.8) with low ROUGE score, indicating that while the vocabulary used in the rationale is different, the overall semantic of the generated rationale remains similar to that of humans. Also, no noticeable changes in the semantic quality of rationale between human transcripts and ASR transcripts because BERTScore is still about 0.8 on both settings.

\input{tables_and_figures/audio_humantranscript}

\noindent \textbf{7. Results on end-to-end audio language models}
We report the results for end-to-end spoken sentiment analysis on PhoWhisper \cite{PhoWhisper} and Qwen2-Audio \cite{Qwen2-Audio}. Based on the results in Table \ref{audio_humantranscript}, we make two observations: First, the performance of PhoWhisper is sub-optimal which we attribute to the fact that it was pre-trained for ASR-based tasks. Second, we found that \textbf{rationale-augmented training can also increase the \sentimentclassification\ performance} for audio language models.

\section{Conclusion}
In this work, we introduce a new task - \sentimentreasoning\   - for both speech and text modalities, along with the framework and  {\textbf{the world's largest multimodal sentiment analysis dataset}}. In \sentimentreasoning, given an input transcript (human transcript or ASR transcript), the model learns to output the sentiment label (POSITIVE, NEUTRAL, or NEGATIVE) and its rationale (the reason for this label). It comprises of two tasks: \sentimentclassification\ and \rationalegeneration.

We menticulously evaluate the use of rationale  during training to improve our models' interpretability and performance. We found that rationale-augmented training improves model performance in \sentimentclassification\ in both human and ASR transcripts (\textbf{  {+2\% increase in both accuracy and macro-F1}}). We found that the generated rationales have different vocabulary to human rationale but with similar semantics. Finally, we found no major difference in the semantic quality of generated rationales between human and ASR transcripts.

\section{Acknowledgement}
We thank Professor Anh Totti Nguyen at Auburn University and Jerry Ngo at MIT for their insightful feedback.

\section{Limitations}
\noindent \textbf{Hybrid ASR}: This study utilized the hybrid ASR system, which is generally recognized as superior in performance compared to the attention-based encoder-decoder or end-to-end ASR systems \cite{luscher19_interspeech, prabhavalkar2023end, raissiluescher}. However, the hybrid ASR requires multiple steps, beginning with acoustic feature extraction and progressing through GMM-HMM modeling before transitioning to DNN-HMM modeling, which complicates reproducibility for non-experts.

\noindent \textbf{Cascaded speech sentiment analysis approach}: While we do report the results for end-to-end systems, our main focus in this paper is on cascaded speech sentiment analysis for \sentimentreasoning\  . This approach uses a previously trained ASR model to generate ASR transcripts that are subsequently input into a language model (LM) for downstream \sentimentclassification\ and \rationalegeneration\ tasks. Consequently, the weights in the ASR model remain unchanged while the LM weights are updated. In this setting, only semantic features from speech are utilized, omitting other trainable acoustic features, like prosody, tones, etc. In spoken language processing, where semantic features play a more important role than other acoustic features, cascaded approach is prefered due to its straightforwardness, simplicity and superior accuracy \cite{lu2023improving, bentivogli-etal-2021-cascade, tran2022does, tseng2023cascading}. Future work should consider the end-to-end sentiment analysis task, where weights in both the ASR model and LM are updated simultaneously, as it might hold promise for improved performance.

\bibliography{custom}

\clearpage 

\appendix

\onecolumn
\tableofcontents
\newpage

\twocolumn

\section{Related Works}
\subsection{Multimodal Speech Sentiment Analysis}
It is widely known that there have been two research directions in the field of speech sentiment analysis, as also confirmed by \citet{switchboard_sentiment}. 
\begin{itemize}
    \item \textbf{Single modality model (unimodal)}: In speech sentiment analysis, single modality models focus on utilizing a single type of data to predict sentiment. These models may rely exclusively on acoustic features, such as pitch, tone, and rhythm, to infer emotional states from spoken language \cite{li2019dilated, li2018attention, wu2019speech, xie2019speech}. Alternatively, they might use raw waveforms \cite{tzirakis2018end, zheng2022two, villatoro2021late} or the textual content of transcripts to predict sentiment \cite{lakomkin2019incorporating}. The strength of single modality models lies in their simplicity and specialization, allowing them to hone in on specific attributes of the data source they are designed for. However, this specialization can also be a limitation, as these models might miss out on the richer, more nuanced information that can be gleaned from combining multiple data types. Despite this, single modality models remain a fundamental approach in the field, providing valuable insights and serving as a benchmark for more complex multimodal systems.
    \item \textbf{Multimodality models}: In speech sentiment analysis, multimodality models leverage the combined strengths of both acoustic and textual data to provide more accurate and nuanced sentiment predictions. While traditional models might rely solely on either the acoustic features—such as tone, pitch, and rhythm—or the text derived from speech transcripts, multimodal models integrate these two data streams. This integration allows for a more holistic understanding of sentiment, as it captures the emotional cues present in the speaker's voice along with the contextual and semantic content of the spoken words. By maximizing the mutual information between these modalities, multimodal models can better discern subtleties in speech that single modality models might miss, leading to accuracy improvements \cite{kim2019dnn, cho2018deep, gu2018multimodal, eskimez2018unsupervised, zhang2019attention}.
\end{itemize}

Our dataset is ideal for both single modal and multimodal research, as it includes both acoustic and text features.
 
\subsection{ASR-based Speech Sentiment Analysis}
Speech sentiment analysis on ASR transcripts is a field that aims to interpret and classify sentiments conveyed in spoken language. As technology advances, ASR systems have become increasingly proficient at transcribing spoken words into text with high accuracy \cite{schneider2019wav2vec, baevski2020wav2vec, baevski2019vq, pmlr-v139-wang21y, Chen2022-qc, Wang2021-zj}, providing a rich source of data for sentiment analysis. Sentiment analysis algorithms then analyze the transcribed text from speech signal, utilizing language models as decoders to detect positive, negative, or neutral sentiments \cite{Lu2020-ke, Shon2021-fi, Wu2022-xl, Tashev2019-jd, Kaushik2017-uc}.

In the era of deep learning, as surveyed by \citet{Al-Qablan2023-mo}, many researchers have been applying deep learning methods to the sentiment analysis process on transcript, leading to the development of various models like Convolutional Neural Networks (CNN), Recurrent Neural Networks (RNN), Long Short-Term Memory (LSTM), and Bidirectional LSTM (BLSTM) \cite{Araque2017-bf, Devipriya2020-dr, Yadav2020-bh}. CNNs, primarily used for image processing, have been adapted for text by treating sentences as sequences of words and applying convolutional filters to capture local features. This approach helps in identifying crucial patterns within the text that are indicative of sentiment \cite{Kumar2020-ih, Wang2020-se}. On the other hand, RNNs are designed to handle sequential data by maintaining a hidden state that captures the history of previous inputs, making them suitable for understanding the context and temporal dependencies in sentences. However, traditional RNNs face challenges with long-term dependencies due to issues like vanishing gradients, which is where LSTMs come in. LSTMs, an advanced form of RNNs, address these issues by incorporating gates that regulate the flow of information, allowing them to maintain and update long-term dependencies effectively. Furthermore, BLSTMs enhance this by processing the input sequence in both forward and backward directions, thus capturing dependencies from both past and future contexts simultaneously. This bidirectional approach is especially useful for sentiment analysis, where the interpretation of a word can depend heavily on both preceding and succeeding words. Together, these architectures provide powerful tools for sentiment analysis, each contributing unique strengths that can be leveraged depending on the specific requirements and characteristics of the data at hand \cite{Gandhi2021-ee, Pal2018-lx, Srinivas2021-mw}.

Developed by Google, BERT (Bidirectional Encoder Representations from Transformers) \cite{Kenton2019-uj} revolutionized NLP tasks by enabling models to understand the context of words in a sentence more effectively through its bidirectional training approach. Unlike previous models that read text input sequentially, BERT reads the entire sequence of words at once, capturing the full context and nuances of language. This capability allows BERT to excel in sentiment analysis, where understanding the subtleties of human emotion and opinion is paramount \cite{Alaparthi2020-ry, Deepa2021-rz}. BERT's pre-training on vast amounts of text data, followed by fine-tuning on specific sentiment analysis tasks, further enhances its performance. By leveraging its powerful language representations, BERT can handle the complexities of sentiment analysis, such as sarcasm, idiomatic expressions, and context-dependent sentiment shifts, making it a preferred choice for applications ranging from social media monitoring to customer feedback analysis. The model's ability to generalize across various domains and languages also contributes to its widespread adoption, offering robust and scalable solutions for sentiment analysis in diverse settings \cite{Hoang2019-it, Xu2019-zf, Sousa2019-mc, Alaparthi2021-lx}.

\subsection{Speech Sentiment Analysis in Healthcare}
Sentiment analysis in healthcare is an emerging field that leverages NLP and machine learning techniques to analyze and interpret the emotional tone conveyed in biomedical textual data. This technology is particularly useful for understanding patient feedback, monitoring public health trends, and improving patient-provider communication. By analyzing large volumes of data from sources such as social media, online reviews, electronic health records (EHRs), and patient surveys, sentiment analysis can provide valuable insights into patient experiences, satisfaction levels, and overall public sentiment towards healthcare services and policies. For instance, analyzing patient reviews on healthcare platforms can help identify common concerns and areas needing improvement, allowing healthcare providers to address issues proactively and enhance the quality of care. Additionally, sentiment analysis can play a critical role in mental health monitoring by detecting signs of distress or dissatisfaction in patient communications, enabling timely intervention and support. As this technology continues to evolve, it holds the promise of transforming healthcare by fostering a more patient-centric approach, enhancing service delivery, and ultimately improving patient outcomes \cite{denecke2015sentiment}. However, the sentiments expressed in clinical narratives have not been extensively analyzed or exploited, based on the total number of previous works we have identified to the best of our knowledge:
\begin{itemize}
    \item Sentiment analysis from the medical web: Most sentiment analysis research in the medical domain focuses on web data, such as medical blogs and forums, to mine patient opinions or assess quality \cite{ali-etal-2013-hear, xia2009improving, na2012sentiment, sokolova2013joe, biyani2013co, ofek2013improving, smith-lee-2012-cross, sharif2014detecting, Patient_rationale}.
    \item Sentiment analysis from biomedical literature: In addition to the analysis of medical social media data, biomedical literature has been examined concerning the outcomes of medical treatments. Within this framework, sentiment denotes the results or efficacy of a treatment or intervention \cite{niu2005analysis, sarker2011outcome}.
    \item Sentiment analysis from medical text (except biomedical literature): Several researchers have focused on leveraging supplementary sources of medical texts to implement sentiment analysis and emotion detection methodologies, suicide notes or patient questionnaire for example \cite{pestian2012sentiment, cambria2012sentic, liu2004conceptnet, cambria2012hourglass}.
\end{itemize}

\textbf{To the best of our knowledge, no literature among those cited has addressed speech sentiment analysis specifically within the domain of healthcare.}



\twocolumn
\section{Details about Data}
\label{sec:Appendix_data}
\subsection{Data Annotation Pipeline}
We use LLM pre-labeling as it helps speed up the labeling process through providing the annotators with the initial sentiment labels and the corresponding rationales. In the relabeling process, annotators go through each sample and inspect it manually. If the annotators deem the label and the rationale is appropriate, they can quickly move to the next sample. If not, the annotators can update the label and rationale to be more appropriate. 

The data annotation process is as followed. First, all the subtitles are separated into different chunks. These segments are subsequently input into gpt-3.5-turbo, which conducts a weakly supervised 3-label classification task to categorize each segment as \textit{NEGATIVE, NEUTRAL, or POSITIVE}. In addition to the sentiment label, gpt-3.5-turbo also provides a brief synthetic rationales for the classification, such as 'Negative medical condition' or 'Objective description'. The labels and rationales generated by gpt-3.5-turbo are subsequently reviewed and independently corrected by a team of 3 developers.

\subsection{LLM Prompt for Pre-labeling}

\begin{tcolorbox}[colback=black!5!white,colframe=black!75!black,title=gpt-3.5-turbo]
\textcolor{blue}{Annotate the sentiment (neutral, positive or negative) of the following sentence and provide a very short justification. The procedure is as follows:}

1. \textcolor{red}{If the segment shows clear emotional signs, annotate based on these signs.}

2. \textcolor{red}{If no emotional markings are present, determine if the segment is an objective description. Positive for beneficial facts/features, negative for detrimental facts/features, and neutral otherwise.}

3. \textcolor{red}{If not objective, check if there's a preference expression. Positive for likes or positive views, negative for dislikes or negative views, and neutral if no preference is expressed.}

4. \textcolor{red}{If too short to determine sentiment, label as neutral.}

\{3 in-context learning examples\}

\end{tcolorbox}

\subsection{Annotation Guidelines}
The definition of "sentiment" encompasses both "emotions" and "facts" in our work. Existing works \cite{switchboard_sentiment, mohammad2016practical, shon2021leveraging, shon2022slue_dataset, shon2023slue} use both emotions and facts for sentiment labeling. 
\begin{itemize}
    \item Emotion: Existing literature includes “emotion” as part of “sentiment” \cite{switchboard_sentiment, shon2021leveraging, mohammad2016practical} and sentiment analysis can be considered a more abstract level of emotion recognition, e.g. polarity of emotions \cite{mohammad2016practical}.
    \item Facts: Many sentiment analysis systems require statements that describe events/situations to be given a sentiment label \cite{switchboard_sentiment, mohammad2016practical}.
\end{itemize}

The annotation task consists of two primary steps. First, annotators are required to perform \sentimentclassification . Second, annotators are instructed to provide a rationale behind each class (\rationalegeneration\ ). 

To ensure consistency, our TESOL-certificated professional linguist has developed an initial guideline inspired by \cite{switchboard_sentiment}, which was also adopted by various well-known works \cite{shon2022slue_dataset, shon2023slue}, and revised it frequently if necessary as followed:

\subsubsection{Output Annotation}
The \textit{NEGATIVE} label is for chunks that discuss negative, serious diseases, disorders, symptoms, risks, negative emotions, or counter-positive statements (e.g. "This would NOT bring a good outcome"). It also applies to incomplete chunks where the amount of negativity is greater than the amount of positivity. 

The \textit{NEUTRAL} label is for incomplete chunks where the ratio of negativity is equal to the ratio of positivity, as well as chunks that describe processes, ask questions, provide advice, or are too short.

The \textit{POSITIVE} label is for chunks that discuss positive outcomes, recovery processes, positive emotions, or counter-negative statements (e.g. "This will \textit{reduce} discrimination"). It also applies to incomplete chunks where the ratio of positivity is greater than the ratio of negativity.

It is important to note that all chunks are considered independent, even though they may be incomplete and related to preceding or following chunks. Given that this data is derived from spoken language, the chunks contain a significant amount of filler words, which are disregarded in the labeling process. The majority of the \textit{NEUTRAL} labels are attributed to chunks that involve sharing advice or descriptions. Additionally, the presence of modal verbs (e.g., should, would, need) often indicates advice sharing, thereby classifying the chunk as \textit{NEUTRAL} regardless of its content.

\subsection{Annotation Flowchart}
Inspired by the well-known annotation flowchart provided by \citet{switchboard_sentiment}, we asked annotators to adopt the annotation flowchart and we ,if necessary, revised as follows:

\begin{enumerate}
    \item Does the segment exhibit distinct emotional cues indicative of sentiment, such as laughter for positive affect or yelling for negative affect?
    \begin{itemize}
        \item \textbf{Yes} – Annotate the corresponding class and also note that: 
        \begin{itemize}
            \item (a) In some instances, individuals may laugh to mitigate the discomfort associated with delivering negative statements. In such cases, it should be classified as neutral.
            \item (b) If individuals exhibit a sneer (a smile or laughter with a mocking tone), the corresponding sentiment should be classified as negative in such instances.
        \end{itemize} 
        \item \textbf{No} - Jump into Step 2
    \end{itemize}

    \item Does the segment provide an objective account of the facts?
    \begin{itemize}
        \item \textbf{Yes} - If the segment lists several positive attributes (e.g., good progress in medical treatment, good signs of health improvement), it is classified as positive. Conversely, if it lists several negative attributes, it is classified as negative. In the absence of a clear preponderance of either, the segment is considered neutral.
        \item \textbf{No} - Jump into Step 3
    \end{itemize}

    \item Does the segment exhibit a preference?
    \begin{itemize}
        \item \textbf{Yes} - If the subjective opinion or preference conveys a like or dislike, or expresses a positive (e.g., "it is beneficial that...") or negative sentiment, it should be annotated accordingly.
        \item \textbf{No} - It's neutral
    \end{itemize}

    \item If the utterance is insufficient in length to accurately assess sentiment, it should be classified as neutral.
\end{enumerate}

\subsection{Data Imbalance Discussion}
As shown in Table \ref{tab:label_distribution}, \textit{NEUTRAL} category is the most predominant, accounting for a significant portion of the dataset. With 3802 instances for both train and test set, \textit{NEUTRAL} sentiments make up approximately half of the dataset. This prevalence of \textit{NEUTRAL} sentiment is expected, as also seen by a real-world conversational dataset \cite{switchboard_sentiment}, given the nature of medical consultations, which often involve objective descriptions, explanations, and advice. The \textit{NEGATIVE} category is the second most common, with around 2395 instances. \textit{NEGATIVE} sentiments include discussions about serious diseases, negative emotions, and adverse medical outcomes. The substantial presence of negative sentiments reflects the medical context, where discussions about illnesses and symptoms are common. The \textit{POSITIVE} category, while the least common, still represents a significant portion of the dataset with 1681 instances. \textit{POSITIVE} sentiments typically involve discussions about recovery processes, positive outcomes, and favorable emotions. 

A slight bias in the distribution of the labels towards NEUTRAL in our dataset (49.94\% in the train set, 43.88\% in the test set) reflects the nature of real-world medical conversations, rather than a weakness of our work. For context, in comparable real-world sentiment analysis datasets such as Switchboard-Sentiment \cite{switchboard_sentiment}, the distribution is as follows: 30.4\% of the speech segments are labelled as POSITIVE, 17\% of the segments are labelled as NEGATIVE, and 52.6\% of the segments are labelled as NEUTRAL. 

To address this labeling bias issue, future works can leverage techniques for fine-tuning models in data imbalance regimes, such as focal loss \cite{ross2017focal}, class weighting \cite{king2001logistic}.

\subsection{Translation Annotation Process and Translation Quality Control}
\label{sec:translation_annotation_quality_control}
The data were initially translated from the source language into target languages (many-to-many) using the Gemini Large Language Model (LLM). Following the annotation process by \citet{zheng2023judging}, the LLM-generated translated transcripts were treated as outputs from a \textit{real} human annotator. In the data quality process, five human annotators manually corrected and then cross-verified \textit{all} these translations based on the context of the whole conversation. Only transcripts that received consensus approval from multiple annotators were retained, resulting in an inter-annotator agreement of 100\%. 

All human annotators possessed a professional language proficiency of C1 or higher (or HSK5 for Chinese) in their respective working languages. Additionally, each annotator had completed basic medical training and demonstrated substantial knowledge of medical terminology in their selected language. Furthermore, they were either currently pursuing or had completed undergraduate or graduate studies in countries where their chosen language is predominantly spoken.

\subsection{Data Samples}
Table \ref{data_samples} shows 9 examples with 3 samples per sentiment label in our dataset. As the Vietnamese transcripts are obtained from short-formed audio, the transcripts contain characteristics of spoken language which serve as noises to the model (e.g. stuttering, hesitation, etc). \textbf{In our English translation, we aim to retain these properties, leading to unnatural, incomplete sentence with broken wording}.

\begin{figure*}
    \centering
    \includegraphics[width=\linewidth]{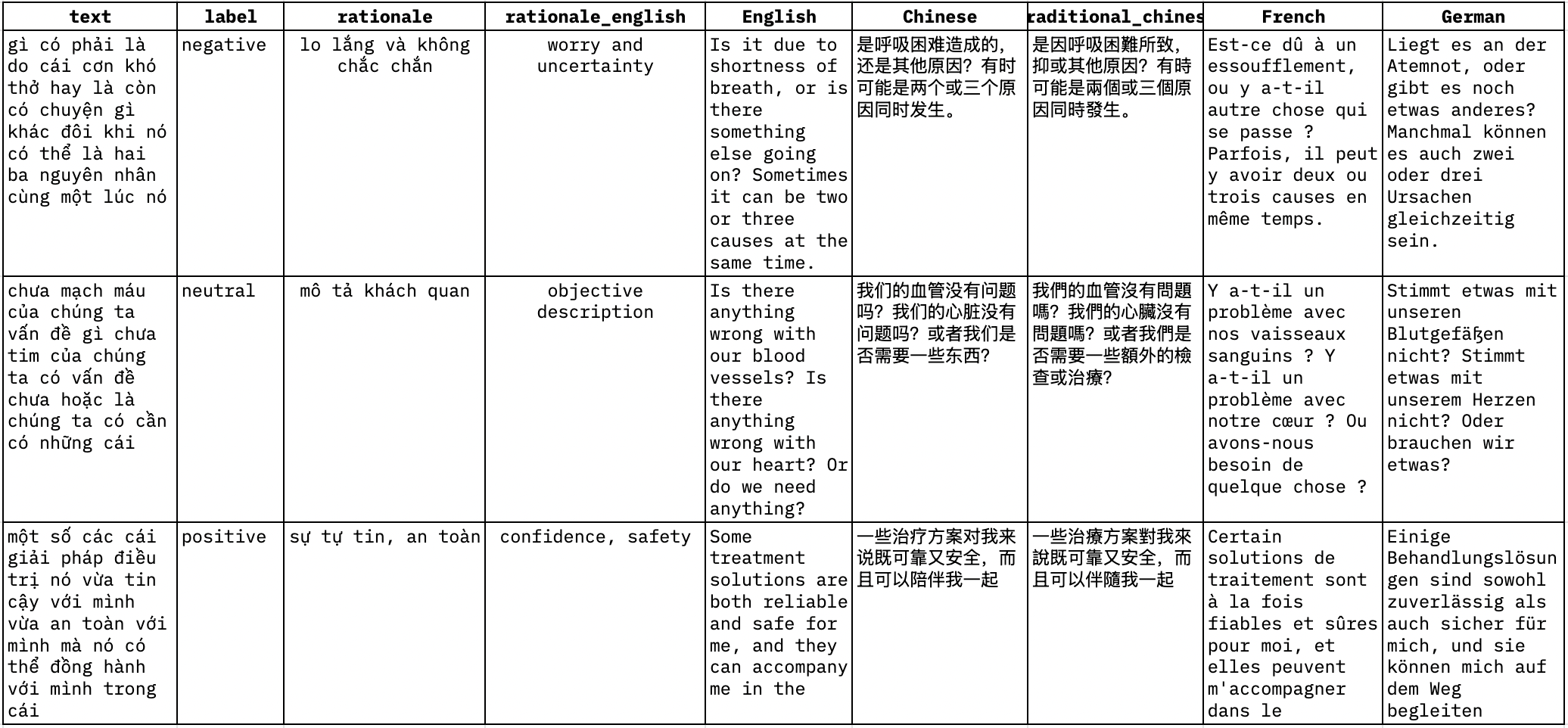}
    \caption{Some samples from our dataset with versions all available languages.}
    \label{fig:other_languages}
\end{figure*}

Figure \ref{fig:other_languages} shows 3 examples per sentiment label for all languages: Vietnamese, English, Chinese (Simplified and Traditional), German and French.

\input{tables_and_figures/data_samples}

\twocolumn
\section{Details about Experimental Setups}
\subsection{Details of ASR Experiments}
\label{sec:details_ASR_experiments}
We employed hybrid ASR setup using wav2vec 2.0 encoder \cite{vietmed_dataset} to transcribe speech to text. First, we generated alignments obtained by using Gaussian-Mixture-Model/Hidden-Markov-Model (GMM/HMM) as labels for wav2vec 2.0 \cite{baevski2020wav2vec} neural network training. The labels used in the acoustic modeling are context-dependent phonemes, triphones in this case. In GMM/HMM process, we used a CART (Classification And Regression Tree) \cite{breiman2017classification} to tie the states $s$, resulting 4501 CART labels: 
\begin{equation}
\begin{split}
&p(x_1^T|w_1^N) = \sum_{[s_1^T]}\prod_{t=1}^Tp(x_t, s_t|s_{t-1}, w_1^N) \\
&= \sum_{[s_1^T]}\prod_{t=1}^T\underbrace{p(s_t|s_{t-1}, w_1^N)}_{\text{transition prob.}}\cdot \underbrace{p(x_t|s_t, s_{t-1}, w_1^N)}_{\text{emission prob.}}    
\end{split}
\end{equation}

After inputting CART labels for hybrid wav2vec 2.0 training, we employed frame-wise cross-entropy (fCE) loss \cite{good1952rational} to train the acoustic model. 

To transcribe speech given the acoustic observations, the acoustic model and n-gram language model \cite{ney1994structuring} should be combined based on the Bayes decision rule using Viterbi algorithm \cite{viterbi} which recursively computes the maximum path to a find best-path in the alignment graph of all possible predicted words to the acoustic observations:
\begin{equation}
\begin{split}
w_1^N &= \operatorname{arg}\max_{N,w_1^N}p\Bigl(\prod_{n=1}^Np(w_n|w_{n-m}^{n-1}) \\
&\cdot \max_{[s_1^T]}\prod_{t=1}^Tp(x_t,s_t|s_{t-1}, w_1^N)\Bigr)    
\end{split}
\end{equation}
Finally, acoustic model and n-gram language model pruning (beam search) is used to only focus on the most promising predicted words at each time step $t$ \cite{ortmanns1997word}.

The final ASR model has 118M trainable parameters and Word-Error-Rate (WER) of 29.6\% on \textit{VietMed} test set.

\subsection{Training Setup}
Our encoders and encoder-decoders were trained on a cluster of 2 NVIDIA A40s with 46 GBs of memory. All models were trained on 30 epochs with with a learning rate of $2e\text{-}5$ and batch size of 64. We evaluated every epoch with early stopping with patience = 3. 

For the decoder-based LLMs, due to their massive number of parameters, we use LoRA \cite{hu2021loralowrankadaptationlarge} for fine-tuning with hyperparameters: the rank of the update matrices $r=8$, and the LoRA scaling factor $\alpha=3$. We train our LLMs for 5 epochs with learning rate 2e-4.

We use the best model checkpoints for evaluation. Note that we do not perform hyperparameter tuning as we only aim to provide the initial benchmark results as well as studying the effects of CoT-augmented finetuning.

\subsection{Student's T-Test}
\begin{figure}[h]
    \centering
    \includegraphics[width=\columnwidth]{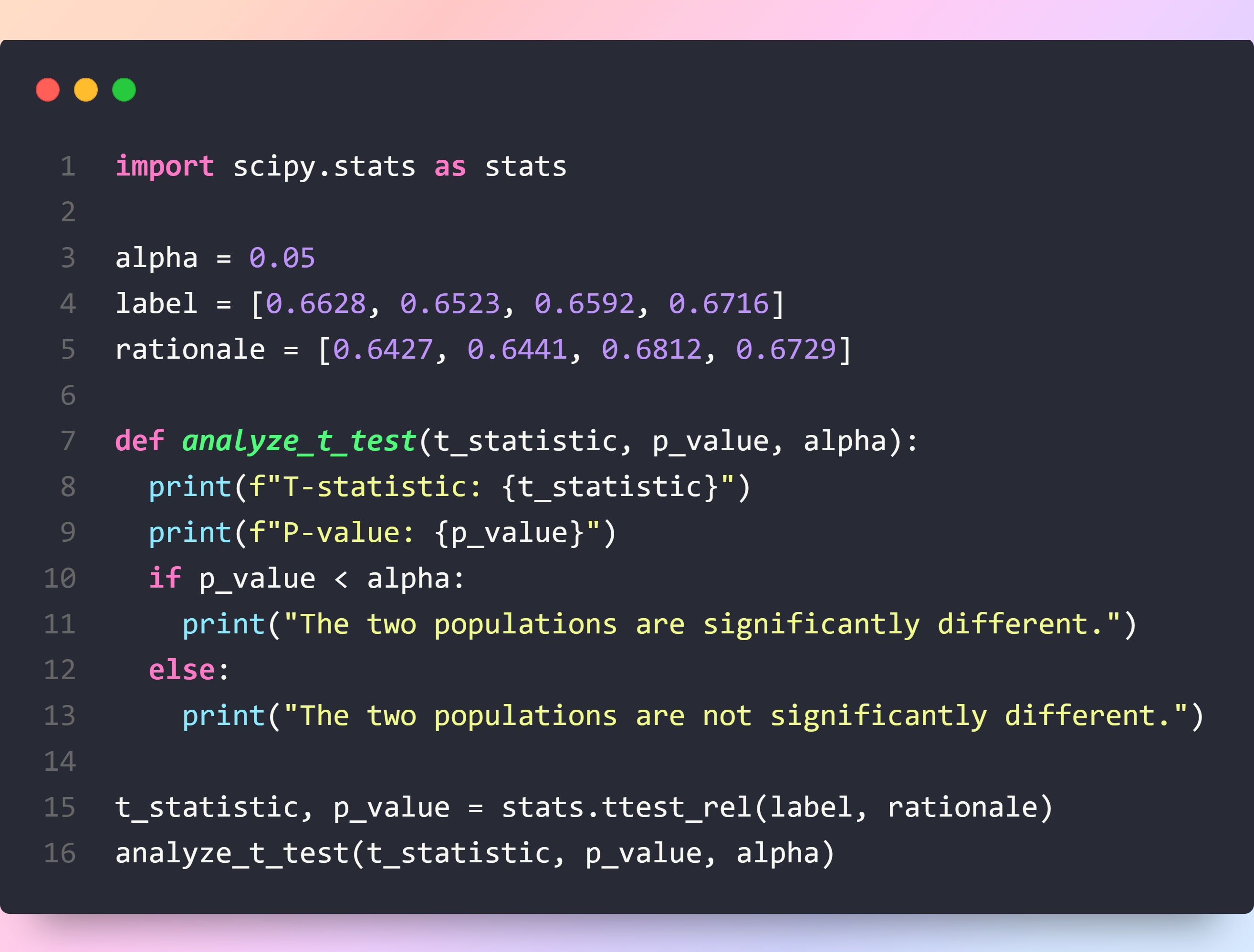}
    \caption{Python code for reproducing Student's t-test experiments}
    \label{fig:t-test_code}
\end{figure}

A Student's t-test, is a statistical method used to compare the means of one or two populations through hypothesis testing. It can assess whether a single group mean differs from a known value (one-sample t-test), compare the means of two independent groups (independent two-sample t-test), or determine if there is a significant difference between paired measurements (paired or dependent samples t-test). Figure \ref{fig:t-test_code} below is the code for reproducing Student's t-test experiments.

\onecolumn
\section{Results on English subset}
\label{sec:results_on_English_subset}
We randomly sampled 50 transcripts and check their quality. We further train English models on this English subset of our dataset to ensure full usability. 

The result of our experiments is in Table \ref{en_humantranscript}. More information on the models used can be found in the same table. Overall, we found that rationale-augmented training also help boost the model's performance. This finding is consistent with what when observed in our experiments in Section \ref{sec:results_and_analysis}.
\input{tables_and_figures/en_humantranscript}

\onecolumn
\section{Error Analysis}
\label{sec:error_analysis}

We report our best model's misclassified transcripts with the highest label confidence (defined as the softmax of the logits of the model prediction) in Table \ref{sample_rationales}. By analyzing at the model's rationale, we hypothesize that the model is confounded by the appearance of certain keywords that elicit either extremely positive ( hữu ích (helpful)) or negative, disease-related words and sentiment which pushes the model away from the \textit{NEUTRAL} label. 

\input{tables_and_figures/sample_rationales}

\end{document}

%% file: tables_and_figures/label_distribution.tex
\begin{table}[h]
  \centering
  \resizebox{0.8\columnwidth}{!}{\begin{tabular}{llrr}
    \hline
    \textbf{Split} &
    \textbf{Label} & \textbf{Count} & \textbf{Percentage} \\
    \hline
    &Neutral & 2844 & 49.94\% \\
Train &Negative & 1694 & 29.74\% \\
    &Positive & 1157 & 20.32\% \\
    \hline
    &Neutral & 958 & 43.88\% \\
Test &Negative & 701 & 32.11\% \\
    &Positive & 524 & 20.01\% \\
    \hline

  \end{tabular}}
  \caption{Distribution of sentiment labels in the dataset for a single language. The real size of the dataset is 6 times larger when accounting all 5 languages - English, Chinese (Simplified and Traditional), German, and French.}
  \label{tab:label_distribution}
\end{table}

%% file: tables_and_figures/datastats_literature.tex
\begin{table*}[h]
\centering
\small
\begin{tabular}{l|c|c|c|c}
\rowcolor[HTML]{F6E4E3} 
\multicolumn{1}{c|}{\cellcolor[HTML]{F6E4E3}\textbf{Dataset}} & \textbf{Venue} & \textbf{\#Samp.} & \textbf{\#Lang.} & \textbf{Domain} \\ \hline
Mosi \cite{zadeh2016mosi}& IEEE & 3k & 1 & Vlog \\
CMU-MOSEI \cite{bagher-zadeh-etal-2018-multimodal} & ACL & 23k & 1 & Various \\
MELD \cite{poria-etal-2019-meld}& ACL & 13k & 1 & TV Series \\
IEMOCAP \cite{busso2008iemocap}& Springer & 12k & 1 & General \\
SEMAINE \cite{5959155}& IEEE & 1k & 1 & Simulation \\ \hline
\sentimentreasoning \ (ours) & - & \textbf{30k} & \textbf{5} & Medical \\ \hline
\end{tabular}%
\caption{Data statistics comparison based on the number of samples and languages. Our dataset with 5 languages (Vietnamese, English, Chinese, German and French) includes 30000 samples, making it \textbf{the largest multimodal sentiment analysis dataset}.}
\label{tab:datastats_literature}
\end{table*}

%% file: tables_and_figures/vi_humantranscript.tex
\begin{table*}[!h]
\centering
\resizebox{\textwidth}{!}{%
\begin{tabular}{lcccccccccc}
\hline
\multicolumn{1}{|l|}{\textbf{Model}} & \multicolumn{1}{c|}{\textbf{Acc.}} & \multicolumn{1}{c|}{\textbf{F1 Neg.}} & \multicolumn{1}{c|}{\textbf{F1 Neu.}} & \multicolumn{1}{c|}{\textbf{F1 Pos.}} & \multicolumn{1}{c|}{\textbf{Mac F1}} & \multicolumn{1}{c|}{\textbf{R-1}} & \multicolumn{1}{c|}{\textbf{R-2}} & \multicolumn{1}{c|}{\textbf{R-L}} & \multicolumn{1}{c|}{\textbf{R-Lsum}} & \multicolumn{1}{c|}{\textbf{BERTscore}} \\ \hline
\multicolumn{11}{c}{\textbf{Encoder (Label Only)}} \\ \hline
\multicolumn{1}{|l|}{PhoBERT} & \multicolumn{1}{c|}{0.6674} & \multicolumn{1}{c|}{0.6969} & \multicolumn{1}{c|}{0.6607} & \multicolumn{1}{c|}{0.6377} & \multicolumn{1}{c|}{0.6651} & \multicolumn{5}{c|}{\multirow{2}{*}{}} \\ \cline{1-6}
\multicolumn{1}{|l|}{ViHealthBERT} & \multicolumn{1}{c|}{0.6752} & \multicolumn{1}{c|}{0.6970} & \multicolumn{1}{c|}{0.6718} & \multicolumn{1}{c|}{0.6535} & \multicolumn{1}{c|}{0.6741} & \multicolumn{5}{c|}{} \\ \hline
\multicolumn{11}{c}{\textbf{Encoder-Decoder (Label Only)}} \\ \hline
\multicolumn{1}{|l|}{ViT5} & \multicolumn{1}{c|}{0.6628} & \multicolumn{1}{c|}{0.6922} & \multicolumn{1}{c|}{0.6687} & \multicolumn{1}{c|}{0.6007} & \multicolumn{1}{c|}{0.6545} & \multicolumn{5}{c|}{\multirow{2}{*}{}} \\ \cline{1-6}
\multicolumn{1}{|l|}{BARTpho} & \multicolumn{1}{c|}{0.6523} & \multicolumn{1}{c|}{0.6870} & \multicolumn{1}{c|}{0.6571} & \multicolumn{1}{c|}{0.5841} & \multicolumn{1}{c|}{0.6427} & \multicolumn{5}{c|}{} \\ \hline
\multicolumn{11}{c}{\textbf{Decoder (Label Only)}} \\ \hline
\multicolumn{1}{|l|}{vmlu-llm} & \multicolumn{1}{c|}{0.6592} & \multicolumn{1}{c|}{0.6768} & \multicolumn{1}{c|}{0.6769} & \multicolumn{1}{c|}{0.5911} & \multicolumn{1}{c|}{0.6483} & \multicolumn{5}{c|}{\multirow{2}{*}{}} \\ \cline{1-6}
\multicolumn{1}{|l|}{Vistral7B} & \multicolumn{1}{c|}{0.6716} & \multicolumn{1}{c|}{0.6858} & \multicolumn{1}{c|}{0.6771} & \multicolumn{1}{c|}{0.6398} & \multicolumn{1}{c|}{0.6676} & \multicolumn{5}{c|}{} \\ \hline
\multicolumn{11}{c}{\textbf{Encoder-Decoder (Label + Rationale)}} \\ \hline
\multicolumn{1}{|l|}{ViT5} & \multicolumn{1}{c|}{0.6633} & \multicolumn{1}{c|}{0.6936} & \multicolumn{1}{c|}{0.6572} & \multicolumn{1}{c|}{0.6335} & \multicolumn{1}{c|}{0.6615} & \multicolumn{1}{c|}{0.3910} & \multicolumn{1}{c|}{0.2668} & \multicolumn{1}{c|}{0.3653} & \multicolumn{1}{c|}{0.3660} & \multicolumn{1}{c|}{0.8093} \\ \hline
\multicolumn{1}{|l|}{BARTpho} & \multicolumn{1}{c|}{0.6619} & \multicolumn{1}{c|}{0.7029} & \multicolumn{1}{c|}{0.6460} & \multicolumn{1}{c|}{0.6265} & \multicolumn{1}{c|}{0.6585} & \multicolumn{1}{c|}{0.3871} & \multicolumn{1}{c|}{0.2613} & \multicolumn{1}{c|}{0.3658} & \multicolumn{1}{c|}{0.3683} & \multicolumn{1}{c|}{0.8077} \\ \hline
\multicolumn{11}{c}{\textbf{Decoder (Label + Rationale)}} \\ \hline
\multicolumn{1}{|l|}{vmlu-llm} & \multicolumn{1}{c|}{0.6729} & \multicolumn{1}{c|}{0.7039} & \multicolumn{1}{c|}{0.6714} & \multicolumn{1}{c|}{0.6307} & \multicolumn{1}{c|}{0.6687} & \multicolumn{1}{c|}{0.3947} & \multicolumn{1}{c|}{0.2467} & \multicolumn{1}{c|}{0.3789} & \multicolumn{1}{c|}{0.3796} & \multicolumn{1}{c|}{0.8086} \\ \hline
\multicolumn{1}{|l|}{Vistral7B} & \multicolumn{1}{c|}{0.6812} & \multicolumn{1}{c|}{0.7152} & \multicolumn{1}{c|}{0.6765} & \multicolumn{1}{c|}{0.6425} & \multicolumn{1}{c|}{0.6781} & \multicolumn{1}{c|}{0.4155} & \multicolumn{1}{c|}{0.2788} & \multicolumn{1}{c|}{0.3880} & \multicolumn{1}{c|}{0.3900} & \multicolumn{1}{c|}{0.8101} \\ \hline
\end{tabular}%
}
\caption{\small{Baseline performance of encoders, encoder-decoders, and decoders on the Vietnamese human transcript. From left to right is: Accuracy, F1-\{negative, neutral, positive, macro\}, ROUGE-\{1, 2, L, Lsum\}, BERTscore. The \textbf{Label Only} models are models trained only with the label, serving as the baseline, while \textbf{Label + Rationale} indicates models trained with rationale. As the \textbf{Label Only} models are not trained to generate rationale, we do not evaluate them on ROUGE and BERTscore.}}
\label{vi_humantranscript}
\end{table*}

%% file: tables_and_figures/vi_ASRtranscript.tex
\begin{table*}[!h]
\centering
\resizebox{\textwidth}{!}{%
\begin{tabular}{lcccccccccc}
\hline
\multicolumn{1}{|c|}{\textbf{Model}} & \multicolumn{1}{c|}{\textbf{Acc.}} & \multicolumn{1}{c|}{\textbf{F1 Neg.}} & \multicolumn{1}{c|}{\textbf{F1 Neu.}} & \multicolumn{1}{c|}{\textbf{F1 Pos.}} & \multicolumn{1}{c|}{\textbf{Mac F1}} & \multicolumn{1}{c|}{\textbf{R-1}} & \multicolumn{1}{c|}{\textbf{R-2}} & \multicolumn{1}{c|}{\textbf{R-L}} & \multicolumn{1}{c|}{\textbf{R-LSum}} & \multicolumn{1}{c|}{\textbf{BERTscore}} \\ \hline
\multicolumn{11}{c}{\textbf{Encoder (Label Only)}} \\ \hline
\multicolumn{1}{|l|}{PhoBERT} & \multicolumn{1}{c|}{0.6166} & \multicolumn{1}{c|}{0.6418} & \multicolumn{1}{c|}{0.6231} & \multicolumn{1}{c|}{0.5658} & \multicolumn{1}{c|}{0.6102} & \multicolumn{5}{c|}{\multirow{2}{*}{}} \\ \cline{1-6}
\multicolumn{1}{|l|}{ViHealthBERT} & \multicolumn{1}{c|}{0.6198} & \multicolumn{1}{c|}{0.6307} & \multicolumn{1}{c|}{0.6261} & \multicolumn{1}{c|}{0.5934} & \multicolumn{1}{c|}{0.6167} & \multicolumn{5}{c|}{} \\ \hline
\multicolumn{11}{c}{\textbf{Encoder-Decoder (Label Only)}} \\ \hline
\multicolumn{1}{|l|}{ViT5} & \multicolumn{1}{c|}{0.6157} & \multicolumn{1}{c|}{0.6412} & \multicolumn{1}{c|}{0.6258} & \multicolumn{1}{c|}{0.5523} & \multicolumn{1}{c|}{0.6064} & \multicolumn{5}{c|}{\multirow{2}{*}{}} \\ \cline{1-6}
\multicolumn{1}{|l|}{BARTpho} & \multicolumn{1}{c|}{0.6056} & \multicolumn{1}{c|}{0.6364} & \multicolumn{1}{c|}{0.6156} & \multicolumn{1}{c|}{0.5311} & \multicolumn{1}{c|}{0.5944} & \multicolumn{5}{c|}{} \\ \hline
\multicolumn{11}{c}{\textbf{Decoder (Label Only)}} \\ \hline
\multicolumn{1}{|l|}{vmlu-llm} & \multicolumn{1}{c|}{0.6216} & \multicolumn{1}{c|}{0.6296} & \multicolumn{1}{c|}{0.6551} & \multicolumn{1}{c|}{0.5186} & \multicolumn{1}{c|}{0.6011} & \multicolumn{5}{c|}{\multirow{2}{*}{}} \\ \cline{1-6}
\multicolumn{1}{|l|}{Vistral7B} & \multicolumn{1}{c|}{0.6299} & \multicolumn{1}{c|}{0.6377} & \multicolumn{1}{c|}{0.6537} & \multicolumn{1}{c|}{0.5609} & \multicolumn{1}{c|}{0.6174} & \multicolumn{5}{c|}{} \\ \hline
\multicolumn{11}{c}{\textbf{Encoder-Decoder (Label + Rationale)}} \\ \hline
\multicolumn{1}{|l|}{ViT5} & \multicolumn{1}{c|}{0.6189} & \multicolumn{1}{c|}{0.6305} & \multicolumn{1}{c|}{0.6286} & \multicolumn{1}{c|}{0.5837} & \multicolumn{1}{c|}{0.6143} & \multicolumn{1}{c|}{0.3571} & \multicolumn{1}{c|}{0.2202} & \multicolumn{1}{c|}{0.3350} & \multicolumn{1}{c|}{0.3366} & \multicolumn{1}{c|}{0.8044} \\ \hline
\multicolumn{1}{|l|}{BARTpho} & \multicolumn{1}{c|}{0.6129} & \multicolumn{1}{c|}{0.6523} & \multicolumn{1}{c|}{0.6028} & \multicolumn{1}{c|}{0.5665} & \multicolumn{1}{c|}{0.6072} & \multicolumn{1}{c|}{0.3956} & \multicolumn{1}{c|}{0.2652} & \multicolumn{1}{c|}{0.3728} & \multicolumn{1}{c|}{0.3774} & \multicolumn{1}{c|}{0.8106} \\ \hline
\multicolumn{11}{c}{\textbf{Decoder (Label + Rationale)}} \\ \hline
\multicolumn{1}{|l|}{vmlu-llm} & \multicolumn{1}{c|}{0.6395} & \multicolumn{1}{c|}{0.6585} & \multicolumn{1}{c|}{0.6557} & \multicolumn{1}{c|}{0.5723} & \multicolumn{1}{c|}{0.6289} & \multicolumn{1}{c|}{0.3853} & \multicolumn{1}{c|}{0.2386} & \multicolumn{1}{c|}{0.3663} & \multicolumn{1}{c|}{0.3671} & \multicolumn{1}{c|}{0.8092} \\ \hline

\multicolumn{1}{|l|}{Vistral7B} & \multicolumn{1}{c|}{0.6354} & \multicolumn{1}{c|}{0.6485} & \multicolumn{1}{c|}{0.6479} & \multicolumn{1}{c|}{0.5892} & \multicolumn{1}{c|}{0.6285} & \multicolumn{1}{c|}{0.3558} & \multicolumn{1}{c|}{0.2237} & \multicolumn{1}{c|}{0.3343} & \multicolumn{1}{c|}{0.3394} & \multicolumn{1}{c|}{0.7994} \\ \hline
\end{tabular}%
}
\caption{\small{Baseline performance of encoders, encoder-decoders, and decoders on the Vietnamese ASR transcript. Further information about our metrics can be found in Table \ref{vi_humantranscript}.}}
\label{vi_ASRtranscript}
\end{table*}

%% file: tables_and_figures/CoT_humantranscript.tex
\begin{table}[!h]
\centering
\setlength{\tabcolsep}{2pt}
\renewcommand{\arraystretch}{1.3}
\resizebox{\columnwidth}{!}{\begin{tabular}{lccccc}
\hline
\multicolumn{1}{|l|}{\textbf{Model}} & \multicolumn{1}{c|}{\textbf{Acc.}} & \multicolumn{1}{c|}{\textbf{F1 Neg.}} & \multicolumn{1}{c|}{\textbf{F1 Neu.}} & \multicolumn{1}{c|}{\textbf{F1 Pos.}} & \multicolumn{1}{c|}{\textbf{Mac F1}} \\ \hline
\multicolumn{6}{c}{\textbf{Encoder-Decoder (Label + Rationale)}} \\ \hline
\multicolumn{1}{|l|}{ViT5\_human} & \multicolumn{1}{c|}{0.6633} & \multicolumn{1}{c|}{0.6936} & \multicolumn{1}{c|}{0.6572} & \multicolumn{1}{c|}{0.6335} & \multicolumn{1}{c|}{0.6615} \\ \hline
\multicolumn{1}{|l|}{ViT5\_elaborate} & \multicolumn{1}{c|}{0.6661} & \multicolumn{1}{c|}{0.6903} & \multicolumn{1}{c|}{0.6799} & \multicolumn{1}{c|}{0.5985} & \multicolumn{1}{c|}{0.6562} \\ \hline
\multicolumn{1}{|l|}{ViT5\_cot} & \multicolumn{1}{c|}{0.6619} & \multicolumn{1}{c|}{0.6968} & \multicolumn{1}{c|}{0.6552} & \multicolumn{1}{c|}{0.6237} & \multicolumn{1}{c|}{0.6586} \\ \hline
\multicolumn{1}{|l|}{BARTpho\_human} & \multicolumn{1}{c|}{0.6619} & \multicolumn{1}{c|}{0.7029} & \multicolumn{1}{c|}{0.6460} & \multicolumn{1}{c|}{0.6265} & \multicolumn{1}{c|}{0.6585} \\ \hline
\multicolumn{1}{|l|}{BARTpho\_elaborate} & \multicolumn{1}{c|}{0.6564} & \multicolumn{1}{c|}{0.7031} & \multicolumn{1}{c|}{0.6528} & \multicolumn{1}{c|}{0.5870} & \multicolumn{1}{c|}{0.6476} \\ \hline
\multicolumn{1}{|l|}{BARTpho\_cot} & \multicolumn{1}{c|}{0.6464} & \multicolumn{1}{c|}{0.6922} & \multicolumn{1}{c|}{0.6611} & \multicolumn{1}{c|}{0.5287} & \multicolumn{1}{c|}{0.6273} \\ \hline
\multicolumn{6}{c}{\textbf{Decoder (Label + Rationale)}} \\ \hline
\multicolumn{1}{|l|}{Vistral7B\_human} & \multicolumn{1}{c|}{0.6812} & \multicolumn{1}{c|}{0.7152} & \multicolumn{1}{c|}{0.6765} & \multicolumn{1}{c|}{0.6425} & \multicolumn{1}{c|}{0.6781} \\ \hline
\multicolumn{1}{|l|}{Vistral7B\_elaborate} & \multicolumn{1}{c|}{0.6688} & \multicolumn{1}{c|}{0.6846} & \multicolumn{1}{c|}{0.6647} & \multicolumn{1}{c|}{0.6564} & \multicolumn{1}{c|}{0.6685} \\ \hline
\multicolumn{1}{|l|}{Vistral7B\_cot} & \multicolumn{1}{c|}{0.6706} & \multicolumn{1}{c|}{0.6725} & \multicolumn{1}{c|}{0.6807} & \multicolumn{1}{c|}{0.6477} & \multicolumn{1}{c|}{0.6670} \\ \hline
\multicolumn{1}{|l|}{vmlu-llm\_human} & \multicolumn{1}{c|}{0.6729} & \multicolumn{1}{c|}{0.7039} & \multicolumn{1}{c|}{0.6714} & \multicolumn{1}{c|}{0.6307} & \multicolumn{1}{c|}{0.6687} \\ \hline
\multicolumn{1}{|l|}{vmlu-llm\_elaborate} & \multicolumn{1}{c|}{0.6867} & \multicolumn{1}{c|}{0.7203} & \multicolumn{1}{c|}{0.6868} & \multicolumn{1}{c|}{0.6353} & \multicolumn{1}{c|}{0.6808} \\ \hline
\multicolumn{1}{|l|}{vmlu-llm\_cot} & \multicolumn{1}{c|}{0.6821} & \multicolumn{1}{c|}{0.6966} & \multicolumn{1}{c|}{0.6779} & \multicolumn{1}{c|}{0.6711} & \multicolumn{1}{c|}{0.6819} \\ \hline
\end{tabular}}%
\caption{\small{Performance of generative models on the different rationale formats on our test set. Human/elaborate/CoT specifies the format of rationale the model was trained on. Details in Section \ref{sec:rationale_format}}}
\label{CoT_humantranscript}
\end{table}

%% file: tables_and_figures/audio_humantranscript.tex
\begin{table}[!h]
\setlength{\tabcolsep}{2pt}
\renewcommand{\arraystretch}{1.3}
\resizebox{\columnwidth}{!}{%
\begin{tabular}{lccccc}
\hline
\multicolumn{1}{|l|}{\textbf{Model}} & \multicolumn{1}{c|}{\textbf{Acc.}} & \multicolumn{1}{c|}{\textbf{F1 Neg.}} & \multicolumn{1}{c|}{\textbf{F1 Neu.}} & \multicolumn{1}{c|}{\textbf{F1 Pos.}} & \multicolumn{1}{c|}{\textbf{Mac F1}} \\ \hline
\multicolumn{1}{|l|}{PhoWhisper} & \multicolumn{1}{c|}{0.4651} & \multicolumn{1}{c|}{0.4393} & \multicolumn{1}{c|}{0.5277} & \multicolumn{1}{c|}{0.3328} & \multicolumn{1}{c|}{0.4333} \\ \hline
\multicolumn{6}{c}{\textbf{Decoder (Label only)}} \\ \hline
\multicolumn{1}{|l|}{Qwen2-Audio } & \multicolumn{1}{c|}{0.5815} & \multicolumn{1}{c|}{0.5707} & \multicolumn{1}{c|}{0.6150} & \multicolumn{1}{c|}{0.5208} & \multicolumn{1}{c|}{0.5688} \\ \hline
\multicolumn{6}{c}{\textbf{Decoder (Label + Rationale)}} \\ \hline
\multicolumn{1}{|l|}{Qwen2-Audio } & \multicolumn{1}{c|}{0.5884} & \multicolumn{1}{c|}{0.5875} & \multicolumn{1}{c|}{0.6131} & \multicolumn{1}{c|}{0.5337} & \multicolumn{1}{c|}{0.5781} \\ \hline
\end{tabular}%
}

\caption{Performance of audio language models}
\label{audio_humantranscript}
\end{table}

%% file: tables_and_figures/data_samples.tex
\begin{table*}[h]
\resizebox{\textwidth}{!}{%
\begin{tabular}{|l|l|l|l|}
\hline
\multicolumn{1}{|c|}{\textbf{Transcript}} & \multicolumn{1}{c|}{\textbf{ENG Translation}} & \multicolumn{1}{c|}{\textbf{Label}} & \multicolumn{1}{c|}{\textbf{Rationale}} \\ \hline
\begin{tabular}[c]{@{}l@{}}bệnh nhân sẽ có những cái rối loạn \\ về mặt cảm xúc đôi khi có những \\ bệnh nhân đã rơi vào trạng thái trầm \\ cảm và đôi khi\end{tabular} & \begin{tabular}[c]{@{}l@{}}The patient will suffer from emotional \\ disorder and sometimes depression\end{tabular} & NEG. & Emotional disorder \\ \hline
\begin{tabular}[c]{@{}l@{}}não đột quỵ đó thì nó liên quan đến \\ việc hình thành các cục máu đông \\ và việc cục máu đông đã nó trôi ra \\ là đi\end{tabular} & \begin{tabular}[c]{@{}l@{}}Stroke is related to the formation of \\ blood clots and the fact that these \\ blood clots travel\end{tabular} & NEG. & \begin{tabular}[c]{@{}l@{}}Negative medical \\ condition\end{tabular} \\ \hline
\begin{tabular}[c]{@{}l@{}}nhầm lẫn với một cái nhóm thuốc \\ khác đó là nhóm thuốc gọi là thuốc\\  chống tiểu cầu tiểu cầu mà cụ\end{tabular} & \begin{tabular}[c]{@{}l@{}}It's often confused with \\ antiplatelet drugs\end{tabular} & NEG. & Confusion \\ \hline
\begin{tabular}[c]{@{}l@{}}điểm cần thiết phải lưu tâm rõ ràng \\ là cái người là bị béo phì đó\end{tabular} & \begin{tabular}[c]{@{}l@{}}A crucial point is that the \\ overweight patient\end{tabular} & NEU. & Sharing advice \\ \hline
\begin{tabular}[c]{@{}l@{}}ra đó là cái hormone cortisol trong\\  máu cũng như là hormone về \\ catecholamine nó\end{tabular} & \begin{tabular}[c]{@{}l@{}}The cortisol hormone in blood as well\\  as catecholamine\end{tabular} & NEU. & \begin{tabular}[c]{@{}l@{}}Objective description \\ of hormones\end{tabular} \\ \hline
\begin{tabular}[c]{@{}l@{}}có thể gọi đây là thuốc lẫn máu \\ hay là một số cái tên khác mà thì \\ nó có thể\end{tabular} & \begin{tabular}[c]{@{}l@{}}You could call these blood-thinning \\ drugs or other names, and it can\end{tabular} & NEU. & Objective description \\ \hline
\begin{tabular}[c]{@{}l@{}}của nó không có cao nhưng mà rất\\  là hình thức thì rất là may mắn là \\ những năm gần đây thì mình có \\ một cái nhóm thuốc khác\end{tabular} & \begin{tabular}[c]{@{}l@{}}It is not expensive, luckily, in recent \\ years there are another group of \\ medicine\end{tabular} & POS. & Expressing luck \\ \hline
\begin{tabular}[c]{@{}l@{}}để mà giảm xóa bỏ cái chuyện \\ hình thành cái cục máu đông đó \\ hiện ta sẽ dùng một số biện pháp\\  trong đó thì chủ\end{tabular} & \begin{tabular}[c]{@{}l@{}}To reduce and eliminate the formation \\ of these blood clots, we use several \\ measures, one of which is\end{tabular} & POS. & \begin{tabular}[c]{@{}l@{}}Avoid forming \\ blood clots\end{tabular} \\ \hline
\begin{tabular}[c]{@{}l@{}}nhóm thuốc này á thì nó là rất là\\  lâu đời và nó không có mất tiền \\ rất là rẻ là\end{tabular} & \begin{tabular}[c]{@{}l@{}}This group of drugs has been around \\ for a very long time and is very \\ cheap, with no cost\end{tabular} & POS. & \begin{tabular}[c]{@{}l@{}}Long-standing and \\ inexpensive medication\end{tabular} \\ \hline
\end{tabular}%
}
\caption{9 examples with 3 samples per sentiment label and its corresponding rationale}
\label{data_samples}
\end{table*}

%% file: tables_and_figures/en_humantranscript.tex
\begin{table*}[!h]
\resizebox{\textwidth}{!}{%
\begin{tabular}{lccccc}
\hline
\multicolumn{1}{|l|}{\textbf{Model}} & \multicolumn{1}{c|}{\textbf{Acc.}} & \multicolumn{1}{c|}{\textbf{F1 Neg.}} & \multicolumn{1}{c|}{\textbf{F1 Neu.}} & \multicolumn{1}{c|}{\textbf{F1 Pos.}} & \multicolumn{1}{c|}{\textbf{Mac F1}} \\ \hline
\multicolumn{6}{c}{\textbf{Encoder (Label Only)}} \\ \hline
\multicolumn{1}{|l|}{mBERT \cite{devlin2018bert}} & \multicolumn{1}{c|}{0.6001} & \multicolumn{1}{c|}{0.5972} & \multicolumn{1}{c|}{0.6320} & \multicolumn{1}{c|}{0.5408} & \multicolumn{1}{c|}{0.5900} \\ \hline
\multicolumn{1}{|l|}{BERT \cite{devlin2018bert}} & \multicolumn{1}{c|}{0.6143} & \multicolumn{1}{c|}{0.6338} & \multicolumn{1}{c|}{0.6245} & \multicolumn{1}{c|}{0.5653} & \multicolumn{1}{c|}{0.6079} \\ \hline
\multicolumn{6}{c}{\textbf{Encoder-Decoder (Label Only)}} \\ \hline
\multicolumn{1}{|l|}{mT0 \cite{muennighoff2022crosslingual}} & \multicolumn{1}{c|}{0.6216} & \multicolumn{1}{c|}{0.6303} & \multicolumn{1}{c|}{0.6418} & \multicolumn{1}{c|}{0.5670} & \multicolumn{1}{c|}{0.6130} \\ \hline
\multicolumn{1}{|l|}{Flan-T5 \cite{https://doi.org/10.48550/arxiv.2210.11416}} & \multicolumn{1}{c|}{0.6157} & \multicolumn{1}{c|}{0.6295} & \multicolumn{1}{c|}{0.6385} & \multicolumn{1}{c|}{0.5462} & \multicolumn{1}{c|}{0.6048} \\ \hline
\multicolumn{6}{c}{\textbf{Encoder-Decoder (Label + Rationale)}} \\ \hline
\multicolumn{1}{|l|}{mT0 \cite{muennighoff2022crosslingual}} & \multicolumn{1}{c|}{0.6175} & \multicolumn{1}{c|}{0.6495} & \multicolumn{1}{c|}{0.6253} & \multicolumn{1}{c|}{0.5535} & \multicolumn{1}{c|}{0.6094} \\ \hline
\multicolumn{1}{|l|}{Flan-T5 \cite{https://doi.org/10.48550/arxiv.2210.11416}} & \multicolumn{1}{c|}{0.6326} & \multicolumn{1}{c|}{0.6487} & \multicolumn{1}{c|}{0.6390} & \multicolumn{1}{c|}{0.5978} & \multicolumn{1}{c|}{0.6285} \\ \hline
\multicolumn{6}{c}{\textbf{Decoder (Label only)}} \\ \hline
\multicolumn{1}{|l|}{Mistral7B \cite{jiang2023mistral7b}} & \multicolumn{1}{c|}{0.6290} & \multicolumn{1}{c|}{0.6536} & \multicolumn{1}{c|}{0.6322} & \multicolumn{1}{c|}{0.5850} & \multicolumn{1}{c|}{0.6236} \\ \hline
\multicolumn{6}{c}{\textbf{Decoder (Label + Rationale)}} \\ \hline
\multicolumn{1}{|l|}{Mistral7B \cite{jiang2023mistral7b}} & \multicolumn{1}{c|}{0.6454} & \multicolumn{1}{c|}{0.6768} & \multicolumn{1}{c|}{0.6364} & \multicolumn{1}{c|}{0.6176} & \multicolumn{1}{c|}{0.6436} \\ \hline
\end{tabular}%
}
\caption{
Baseline performance of encoders, encoder-decoders, LLMs on the English human transcript. Further information about our metrics can be found in Table 2.}
\label{en_humantranscript}
\end{table*}

%% file: tables_and_figures/sample_rationales.tex
\begin{table*}[h]
\resizebox{\textwidth}{!}{%
\begin{tabular}{|l|c|c|l|l|}
\hline
\textbf{Transcript} & \textbf{Label} & \textbf{Pred.} & \textbf{Human Rationale} & \textbf{Model Rationale} \\ \hline
\begin{tabular}[c]{@{}l@{}}\textbf{VI:} trả lại cho họ chất lượng cuộc \\ sống bình thường như bao \\ người khác là được nghe được\\ nói thế nhưng điều kỳ diệu đã \\ \textbf{ENG:} give them back a normal \\ quality of life like everyone \\ else, but a miracle has \\ happened\end{tabular} & NEU. & POS. & \begin{tabular}[c]{@{}l@{}}Mô tả khách quan\\ (Objective \\ description)\end{tabular} & \begin{tabular}[c]{@{}l@{}}chất lượng cuộc sống \\ bình thường \\ (normal quality of life)\end{tabular} \\ \hline
\begin{tabular}[c]{@{}l@{}}\textbf{VI:} những chia sẻ vô cùng hữu \\ ích và thiết thực vừa rồi ạ \\ có thể thấy là hầu hết người \\ bệnh nằm điều trị trong \\ \textbf{ENG:} with the extremely useful \\ and practical shares shared\\ just now, it can be seen that\\ most of the patients are in \\ hospital for treatment)\end{tabular} & NEU. & POS. & \begin{tabular}[c]{@{}l@{}}Mô tả khách quan\\ (Objective \\ description\end{tabular} & \begin{tabular}[c]{@{}l@{}}thông tin hữu ích và \\ thiết thực\\ (useful and practical\\ information)\end{tabular} \\ \hline
\begin{tabular}[c]{@{}l@{}}\textbf{VI:} khám suốt tiểu đường nó \\ vẫn mệt mỏi vô khám tai \\ biến bộ não vô khám nhưng\\ mà xương thì nó loãng \\ xương rất là nhiều\\ \textbf{ENG:} even after being examined\\ for diabetes, she still feels \\ tired, has had a stroke, and \\ has not been examined for \\ stroke, but her bones have \\ a lot of osteoporosis\end{tabular} & NEU. & NEG. & \begin{tabular}[c]{@{}l@{}}Mối quan tâm và \\ vấn đề sức khỏe\\ (Health concerns \\ and problems)\end{tabular} & \begin{tabular}[c]{@{}l@{}}triệu chứng tiêu cực \\ của bệnh tiểu đường\\ và loãng xương \\ (negative \\ symptoms of \\ diabetes and \\ osteoporosis)\end{tabular} \\ \hline
\end{tabular}%
}
\caption{Some misclassified transcripts from our best model with high confidence (>0.99). VI means the Vietnamese transcript, EN means the transcript translated to English}
\label{sample_rationales}
\end{table*}